\documentclass{article}

\usepackage{microtype}
\usepackage{graphicx}
\usepackage{subcaption}
\usepackage{booktabs}
\usepackage{hyperref}

\usepackage[preprint]{icml2026}

\usepackage{amsmath}
\usepackage{amssymb}
\usepackage{mathtools}
\usepackage{amsthm}
\usepackage[capitalize,noabbrev]{cleveref}

\theoremstyle{plain}

\theoremstyle{definition}

\newcommand{\grind}{\texttt{grind}}
\newcommand{\ematch}{\texttt{e-match}}

\newcommand{\githubrepo}{https://github.com/aurasoph/NeuralGrind}

\icmltitlerunning{Learned Interventions Inside Lean 4's grind}

\begin{document}

\twocolumn[
\icmltitle{Learned Interventions Inside Lean~4's \texttt{grind}}

\icmlsetsymbol{equal}{*}

\begin{icmlauthorlist}
\icmlauthor{Evan Wang}{yyy,cse,math}
\icmlauthor{Simon Chess}{equal,yyy,cse,math}
\icmlauthor{Sophie Szeto}{equal,yyy,cse,math}
\icmlauthor{Theodore Meek}{equal,yyy,math}
\end{icmlauthorlist}

\icmlaffiliation{yyy}{Math AI Lab, University of Washington, Seattle, United States}
\icmlaffiliation{cse}{Department of Computer Science and Engineering, University of Washington, Seattle, United States}
\icmlaffiliation{math}{Department of Mathematics, University of Washington, Seattle, United States}

\icmlcorrespondingauthor{Evan Wang}{aurasoph@uw.edu}

\icmlkeywords{Lean 4, grind, premise selection, lookahead, theorem proving, machine learning}

\vskip 0.3in
]

\printAffiliationsAndNotice{\icmlEqualContribution}

\begin{abstract}
Lean~4's \grind{} tactic combines congruence closure, \ematch{}ing, and case-splitting into a single automated solver, and like any such solver, it relies on hand-tuned heuristics to decide what to instantiate and where to case-split. These heuristics are tempting targets for learning, but there is a catch: because \grind{}'s search is non-monotone, a learned heuristic that helps one proof can break another, and an always-on replacement usually nets out near zero. We avoid this by invoking a learned intervention only after stock \grind{} has already failed: a failure-triggered cascade that, by construction, cannot lose a proof \grind{} already had. We apply it to two of \grind{}'s internal decisions. A cost-aware \ematch{} filter solves slightly more problems and runs about 5\% faster. A lookahead step proves five theorems it otherwise times out on. We also report the negative result that motivated the design: across four feature-based models, statically predicting the correct case split is no better than random, because whether a split explodes is a runtime property that the features do not capture. Our results suggest that learning within theorem-proving tactics is most effective as a mechanism for deciding when and how to spend bounded search, backed by a reliable symbolic fallback.\footnote{Code: \href{\githubrepo}{GitHub repository}.}
\end{abstract}

\section{Introduction}
Automated theorem provers increasingly rely on a small number of powerful tactics as their reasoning backends. This makes the internal search decisions of those tactics a high-leverage target for learning: improving a single branching, instantiation, or pruning heuristic can affect many downstream proof attempts. At the same time, these decisions are difficult to learn safely. A heuristic that appears better locally may change the shape of the symbolic search enough to create new blow-ups, causing regressions on goals that the original tactic already solved.

We study this problem within \grind{}~\citep{lean_grind}, a recent tactic for the Lean~4 proof assistant~\citep{demoura2021lean4}. It integrates SMT-style automation~\citep{demoura2008z3} into Lean by combining congruence closure, \ematch{}-based lemma instantiation, and case splitting in a single procedure. Many of its decisions are heuristic: which \ematch{} instances to keep, which goal to split next, and in what order to explore the resulting cases. A single proof may involve thousands of such choices, which makes \grind{} a natural place to ask whether learned guidance can help.

Most machine-learning work for interactive theorem proving operates at the level of tactics or proof steps, outlined in \cref{sec:related}. We instead look inside one tactic and modify some of its internal choices rather than replacing the tactic itself. This keeps the original search procedure in the loop: stock \grind{} provides both a baseline and a fallback.

The main complication is that \grind{}'s search is non-monotone. \ematch{}ing can grow very quickly, so a choice that appears better in the short term may make the overall search worse. In our experiments, always-on learned replacements have this character: they solve some goals that stock \grind{} misses, but also break goals that stock \grind{} can already prove. We therefore use learning more conservatively. Rather than trying to predict the right choice from static features, we run a cheap lookahead that is triggered only after stock \grind{} fails. This gives a failure-triggered cascade: stock \grind{} is tried first, and the intervention is applied only to the goals it leaves open. As a result, the intervention does not endanger proofs that \grind{} already solves.

We evaluate this design with working Lean code. Our implementation includes a cost-aware \ematch{} filter and a lookahead step that proves some theorems beyond stock \grind{}. We also report a negative result: a static model for predicting the next split performs no better than random choice. The lesson, then, is not that a neural model should replace \grind{}'s heuristics, but that learning is best used to locate where static choices fail and to steer bounded search behind a symbolic fallback.

Our main contributions are:
\begin{enumerate}
    \item We formulate failure-triggered learned intervention as a safe deployment pattern for learning inside non-monotone theorem-proving tactics.
    \item We implement two Lean-native interventions inside \grind{}: a cost-aware \ematch{} filter and a bounded lookahead procedure for case splitting.
    \item We show that these interventions improve theorem-level performance without sacrificing baseline solves: the \ematch{} filter gives a small speed and success gain on 855 held-out theorems, while the lookahead cascade rescues five stock \grind{} timeouts with zero regressions.
    \item We show that static feature prediction is insufficient for the rescuable split decisions: four learned policies fail to beat random choice precisely where it matters, suggesting that branch explosion is primarily a dynamic property of the search.
\end{enumerate}

\section{Related Work}
\label{sec:related}
\paragraph{Learned heuristics in SAT and SMT.}
The choices we study---which goal to branch on and which instantiations to keep---are close to the branching, restart, and instantiation heuristics used in SAT and SMT solvers. There is a long line of work on learning such heuristics. NeuroSAT~\citep{selsam2019neurosat} learns a solver from single-bit supervision; Graph-Q-SAT~\citep{kurin2020graphqsat} uses a graph network as a branching policy for CDCL; and FastSMT~\citep{balunovic2018fastsmt} learns to compose Z3 tactic strategies. These systems score the current state from its features. Our split-rescue experiments suggest that this is not enough for the choices inside \grind{}: on the high-impact decisions, static scoring is no better than random choice (\cref{tab:rescue}). The issue is that the cost of a branch is often visible only after taking it. Our positive result therefore uses lookahead rather than a purely static prediction.

\paragraph{E-graphs and equality saturation.}
\grind{} maintains an e-graph through congruence closure, the same basic data structure used by equality-saturation systems such as egg~\citep{willsey2021egg}. In those systems, a central problem is controlling e-graph growth by deciding when to apply rewrite rules. Our \ematch{} filter addresses a similar problem inside \grind{}: it tries to discard low-value instantiations before they are added to the e-graph. The failures of this filter are also useful. The heavy-tailed, lemma-diverse explosions described in \cref{app:filter-neg} show why simple pruning rules are not enough to control growth in all cases.

\paragraph{Learning for interactive theorem proving.}
Most learning-based work in interactive theorem proving operates at the level of tactics or proof steps. Examples include language models that generate proof steps~\citep{polu2020gptf,yang2023leandojo}, learned tactic policies~\citep{gauthier2021tactictoe}, and neural-guided proof search~\citep{lample2022htps,silver2018alphazero}. Premise selection and hammer systems instead retrieve useful lemmas for an external prover~\citep{alemi2016deepmath,blanchette2016hammering,czajka2018coqhammer}. Our setting is different: we keep the tactic fixed, and learn only some of its internal choices. This keeps the original symbolic search as the baseline and fallback.

\paragraph{Algorithm selection and portfolios.}
Our failure-triggered cascade is related to algorithm selection and portfolio solvers such as SATzilla~\citep{xu2008satzilla}. Portfolio methods usually pick a solver or configuration up front, based on features of the instance. Our cascade makes the decision later. It first runs stock \grind{}, and invokes the learned procedure only on goals that remain unsolved. This avoids having to predict in advance whether learning will help, which is exactly the prediction problem that static features handle poorly in our experiments.

\section{Integrating Learning Into \grind{}}
\label{sec:system}
\grind{} runs an action loop, which can be viewed schematically as
\[
\texttt{solvers} \;\triangleright\; \texttt{instantiate}
   \;\triangleright\; \texttt{splitNext} \;\triangleright\; \texttt{mbtc}.
\]
The loop repeats until it reaches a fixpoint or runs out of heartbeats. On each pass, the subsolvers propagate known facts, instantiate lemmas, split goals, or close branches.

We use three places where learned guidance can be inserted without forking the tactic: (i) the \ematch{} instance filter, which can discard low-value instantiations before they enter the congruence structure; (ii) split-target selection, implemented by \texttt{splitNext}; and (iii) premise augmentation, which chooses facts to assert before the search begins. Everything is Lean-native with sub-millisecond latency. We summarize the data, features, and held-out evaluation protocol in \cref{app:method}.

The learned components are deliberately small. In this setting, where the model is applied matters more than how large it is, and our scaling results in \cref{app:filter-neg} support this. The \ematch{} filter is a binary classifier that scores each candidate instantiation for proof-relevance---whether it will appear in the final proof term. It uses features of the lemma being instantiated, including its identity, head symbol, conclusion tokens, and premise tokens; matching features of the current goal; and a few numeric signals, such as the \ematch{}ing round and how often the lemma has been useful before. These form a $135$-dimensional vector fed to a three-layer MLP ($135\!\to\!64\!\to\!32\!\to\!1$) trained with binary cross-entropy. The split-study models are similarly small: MLPs and gradient-boosted trees over goal- and candidate-level features (\cref{app:split-models}).

\section{Improvement 1: A Cost-Aware \ematch{} Filter}
\label{sec:filter}
\ematch{}ing~\citep{demoura2007ematching} instantiates quantified lemmas against the current e-graph, which is maintained by congruence closure~\citep{nelson1980congruence}. Many instantiations do not help the proof, but they still cost work--each match adds facts or terms that later solver components must maintain, and a single quantified lemma may match in many ways.

We train a lightweight classifier on per-instance features and use it at the \ematch{} call site to drop low-value instantiations. On a held-out suite of $855$ theorems, the filter runs about $5\%$ faster, and additionally recovers $+2$ solves over stock \grind{}. Scaling training data by $20\times$ does not broaden the niche, suggesting a limitation of the lemma-identity mechanism rather than a shortage of examples. We therefore treat the filter as a useful speed component and evidence for where static scoring is insufficient.

\section{Improvement 2: Lookahead for Splitting}
\label{sec:split}
When \ematch{} and congruence closure stop making progress, \grind{} falls back on case splitting. It chooses a fact and branches on its possible cases, for example, $x=0$ versus $x\neq0$, then continues the search separately on each branch. The split target is chosen from a pool of candidates using a fixed numeric tie-breaker. A good split may close the goal quickly, while a bad split may create branches that grow until the tactic times out. 

To study this choice, we use an oracle experiment. At each decision with multiple candidates, we force each candidate in turn, rerun \grind{} with the rest of the proof fixed, and record the outcome. Each forced run is an actual execution of the policy that chooses that candidate. In total, we collect about $16$K forced-choice outcomes from $4{,}120$ multi-candidate decisions over \textsc{numina}.

\paragraph{The main benefit addresses capability instead of speed.} An oracle that always chooses the cheapest split reduces the total number of splits by only about $4\%$. The more interesting cases are failures of stock \grind{}. In $675$ decisions, \grind{}'s chosen split leads to a timeout. In $97$ of those decisions ($14\%$), at least one other split available at the same point closes the goal without new lemmas or extra search. We call these the rescuable failures: they are the decisions where simply picking a different available candidate would have rescued the proof.

For example, in a rational equation such as $1/(x{+}1)+1/(x{+}2)=1/x$, \grind{} may split on a fact that sends one branch into a blow-up and eventually times out. A different available split would close the goal after a few more steps. On these rescuable decisions, stock \grind{} has a $0\%$ rescue rate by construction, while a uniformly random alternative succeeds $57\%$ of the time; see \cref{tab:rescue}. This leaves room for a better split policy.

\paragraph{Static prediction does not beat random on the rescuable failures.} We tried four policies for choosing the rescuing split: a gradient-boosted cost model~\citep{chen2016xgboost}, a generation-ordering rule validated on a held-out cost metric, a doom/value model, and a failure-aware success classifier with candidate AUC 0.85 on by-decision held-out data. None beats random on the rescuable decisions; see \cref{tab:rescue}. These models have strong aggregate metrics overall and presumably help on the easy or median decisions, but precisely on the decisions where \grind{}'s own heuristic was wrong, the static models' predictions are correlated enough with \grind{}'s signal that they tend to be wrong in the same way. The missing information is whether a split will cause a branch to explode, which appears to be a dynamic property rather than one captured by our static features, outlined in \cref{app:split-models}.

\begin{table}[ht]
\centering\small
\caption{Static-feature policies do not beat random on the rescuable
split failures. Lookahead succeeds because it tries the split instead of
predicting its outcome from static features.}
\begin{tabular}{lr}
\toprule
Policy (on the rescuable split-failures) & rescue rate \\
\midrule
\grind{} (stock heuristic)               & 0\% \\
GBM cost model / generation rule         & 30--46\% \\
value-doom / failure-aware classifier    & 46--53\% \\
\textbf{uniform random}                  & \textbf{57\%} \\
\midrule
\emph{1-step lookahead (realized)}       & \emph{100\%} \\
\bottomrule
\end{tabular}
\label{tab:rescue}
\end{table}

Lookahead captures most of the benefit by turning split selection into a small experiment. For each candidate, the policy makes a throwaway copy of the goal, forces that split, and runs \grind{} on the copy for a bounded number of steps. We say a trial closes when this bounded sub-search proves the goal in the copy; the parent tactic commits to the first such candidate and discards the other copies. Since the trial actually executes the split, it observes the dynamic behavior that the static models miss. The question is therefore not whether lookahead can identify the rescuing split, but how much it costs to try.

On the rescue decisions, a time-capped trial recovers $90\%$ of rescuable failures with a $15$\,s cap. Even when every candidate is tried, the average trial cost remains below the time \grind{} would otherwise spend following its bad choice to timeout (\cref{tab:cap}).

\begin{table}[ht]
\centering\small
\caption{Time-capped lookahead on the rescue decisions. At $C{=}15$\,s it
rescues $90\%$ of the rescuable failures, at lower cost than the
approximately $53$\,s \grind{} spends on its bad choice.}
\begin{tabular}{lcccc}
\toprule
cap $C$ (s)                 & 5 & 10 & \textbf{15} & 20 \\
\midrule
rescued (of 97)             & 0\% & 80\% & \textbf{90\%} & 92\% \\
cost (s/decision)           & 21  & 39   & 45            & 50  \\
\bottomrule
\end{tabular}
\label{tab:cap}
\end{table}

\paragraph{Proving theorems stock \grind{} misses.}
We implemented the lookahead split policy inside \grind{}. At each split,
the tactic tries candidates on functionally discarded goal copies under
an iteration budget, and commits to a candidate whose trial closes.

We then ran the implementation live. On the $44$ theorems whose traces
contain a rescue decision, stock \grind{} solves $39$ and times out on
$5$ under a deterministic $400$K-heartbeat budget. Always-on lookahead
solves $42$: it rescues all
$5$ stock failures, but regresses on $2$ goals that stock \grind{} already
solves. These regressions come from the greedy rule of taking the first candidate whose trial closes, which is not always globally best.

This motivates the cascade deployment. We first run stock \grind{}, and
run lookahead only if stock \grind{} fails. The $39$ stock solves are then
left unchanged, so the $2$ always-on regressions do not occur, while the
$5$ stock failures are rescued. On this set, the cascade solves $44/44$
theorems instead of $39/44$, a net gain of $5$ with no regressions, and
uses fewer total splits ($1059$ versus $1271$ at the 400K budget).

\paragraph{When lookahead works.}
Lookahead is a shallow probe: it rescues goals when the right split closes within the trial budget, but it is not a substitute for deeper search. The $5$ rescued theorems have shallow rescues: after the right split, the branch closes within roughly $9$ further splits. The $2$ regressions require deeper, multi step commitments, with rescuing branches between $10$ and $57$ splits deep, sometimes spread across several sequential decisions. In those cases, stock \grind{}'s ordinary unbounded search make the better long-term choice. Thus, the cascade is safer than enabling lookahead globally.

\begin{table}[ht]
\centering\small
\setlength{\tabcolsep}{3pt}
\caption{On rescuable decisions, dynamic execution is the lever. A non-learning probe that tries candidates in random order until one closes rescues nearly all of them ($\approx$$96\%$, the remainder hitting the per-decision trial cap) in ${\sim}1.6$ candidate trials on average, while the learned static policies of \cref{tab:rescue} fall below even the random single pick. ``$\le 4.6$ trials'' for the deployed lookahead is the average number of candidate trials it spends per rescue decision when run without ordering, bounded above by the per-decision candidate count. There is room for learning is in cost and reach (when and how deep to probe), not in naming the rescuing split.}
\begin{tabular}{lccc}
\toprule
policy & dynamic? & rescue rate & trials to rescue \\
\midrule
stock \grind{}            & no  & 0\%           & --   \\
random single pick        & no  & 57\%          & 1    \\
random bounded probe      & yes & ${\sim}100\%$ & 1.6  \\
lookahead (try all)       & yes & 100\%         & $\le 4.6$ \\
\emph{perfect order}      & --  & 100\%         & 1.0  \\
\bottomrule
\end{tabular}
\label{tab:baselines}
\end{table}

\section{Discussion: The Cascade Discipline}
\label{sec:discussion}
The trouble with putting a learned heuristic inside \grind{} is that the tactic's search can blow up: \ematch{}ing keeps generating new facts, so a choice that looks good at one step can leave the solver with far more work a few steps later. A learned rule that is switched on all the time runs straight into this. It steers some proofs to a quick close and sends others into blow-up, and across a benchmark, the two can roughly cancel.

Our fix is to never let the learned rule touch a proof that already works. We run stock \grind{} first, and call the learned intervention only on the goals where stock \grind{} has failed. The lookahead applies this idea at a single split: try the alternatives, and keep one if it closes. The filter is an easier version of it, dropping an instantiation only when it is confident the instantiation is useless. The same recipe also helps with premise augmentation: if we add premises only as a retry after \grind{} fails, we keep the rescues and lose nothing (\cref{app:premise}).

This changes where we think learning should go. It should be emphasized that the deployed split policy itself is non-learned: it is a greedy first-closing probe, and \cref{tab:baselines} shows that a non-learning random-order bounded probe captures most of the benefit. Learning is more suitable one layer up: deciding when lookahead is worth its cost, and which candidates to try first so the probe finds a closing trial in fewer attempts. \cref{tab:rescue} shows that a form of this, picking the right split from static features, does not work in the most impactful cases. Four different models do about as well as random, likely because the most impactful features depends on what happens next, which the models struggle to learn. What works is executing the split under a bounded budget and observing its behavior. The job of a learned model is therefore not to name the split, but to help steer the search: a gating policy that decides when the probe is worth the time, and a candidate-ordering policy that reduces the average probe cost.

\section{Conclusion}
We integrated learning into Lean~4's \grind{} at several internal call sites and found two problem-level improvements. A cost-aware \ematch{} filter gives a small success and speed improvement on a held-out suite, while a bounded lookahead cascade proves five theorems stock \grind{} cannot, with zero regressions. The results support a simple rule: preserve the symbolic solver as the default, and use learning to decide when and how to spend bounded search on the failures it leaves behind.

Two directions follow directly from the analysis here. First, the discussion identifies a concrete role for learning that we did not build: a gating model that decides when lookahead is worth running, and an ordering model that reduces the average candidate trials per rescue toward the perfect-order baseline of 1. Second, premise augmentation is a third natural intervention surface (\cref{app:premise}); the patterns that protects the lookahead intervention from regression should extend to a learned premise retriever.

\section*{Impact Statement}
This work studies machine-learning interventions inside an automated
reasoning tactic for the Lean~4 proof assistant. The aim is to make formal
verification more capable; we see no specific societal risks beyond those
general to automated theorem proving.

\section*{Acknowledgements}
We thank the UW Department of Applied Mathematics and the organizers of the University of Washington 2026 Lean Hackathon for access to compute.

\bibliography{neuralgrind}
\bibliographystyle{icml2026}

\appendix
\onecolumn

\section{Models, Data, and Held-out Methodology}
\label{app:method}
\textbf{Held-out invariant.} Two suites are held strictly out of all
training: a $75$-theorem active-split benchmark and an $855$-theorem
held-out suite (\texttt{held\_out\_v2}); a programmatic gate enforces that
no training row originates from either. \textbf{Forced-choice protocol.}
For split analysis we instrument the trace observer to log, at each
multi-candidate decision, the candidate \grind{} would pick; we then
re-run \grind{} forcing each alternative candidate with the remainder of
the proof fixed and record outcome and split count. Because a forced run
is a realized deployment of the policy that picks that candidate, the
resulting rescue rates are measured, not projected. \textbf{Inference.}
All learned scorers are exported to a native format read inside Lean, so
scoring adds sub-millisecond latency per decision.

\paragraph{Filter model and training.}
The filter is a concatenation MLP with no cross-features. Each
instantiation is represented by eight $16$-dimensional embeddings---the
lemma identity, its tokenized name, its head symbol, and pooled tokens of
its conclusion and premises, together with the goal's head symbol and
pooled conclusion and premise tokens---plus seven numeric features (the
\ematch{}ing round, the instance's ordinal within the round, the number of
new instances in the round, per-theorem instance and round counts, and the
lemma's historical usefulness frequency and positive rate). The resulting
$135$-dimensional vector passes through
$\mathrm{Linear}(135,64)\!\to\!\mathrm{ReLU}\!\to\!\mathrm{Dropout}(0.1)
\!\to\!\mathrm{Linear}(64,32)\!\to\!\mathrm{ReLU}\!\to\!\mathrm{Linear}(32,1)$,
trained with binary cross-entropy (Adam) to predict proof-relevance: the
label is $1$ when the instance's $(\text{uid},\text{lemma})$ pair appears
in the closing proof term, recovered from a \texttt{proof\_relevant} trace
event. The model is small by design---the shipped version trains on
${\sim}1.2$K Mathlib-derived rows---and the scaling study
(\cref{app:filter-neg}) confirms that capacity, not the architecture, is
the limit. At inference the weights are serialized to a fixed-layout
binary and scored by an evaluator compiled into the tactic, well under a
millisecond per instance; an instance is dropped when its score falls
below a threshold $\tau$, subject to a per-round drop ratio.

\section{Filter: Negative and Niche Results}
\label{app:filter-neg}
\textbf{Data scaling plateaus.} Increasing augmented training rows by
$20\times$ ($\sim$3.9K to $\sim$84K, sweeping the full eligible
\textsc{finelean}+\textsc{numina}+\textsc{workbook} pool) leaves held-out
success \emph{exactly} at $830/855$ across configurations---zero marginal
benefit, and $-3$ relative to our shipped Mathlib-trained filter (the
$833/855$ of \cref{sec:filter}; stock \grind{} solves $831$).
\textbf{Lemma vocabulary is not the bottleneck.} Inline trace-time
feature capture grows the lemma vocabulary $3.5\times$ ($73\to257$) and
lifts test AUC $0.949\to0.955$, but held-out gains only $+1$
($829\to830$). \textbf{Threshold tuning does not bridge the gap.} Sweeping
$(\tau,\text{drop\_ratio})$ over five settings holds success flat at
$829$--$830$; none reaches the shipped filter's $833$. \textbf{Balancing helps
once, then hurts.} A balanced $\approx\!32/32/2/34\%$ mix scores $832$
with fewer rows, but doubling the data while up-weighting one source
($14\times$ more \textsc{workbook}) regresses to $829$: distribution
balancing is necessary but not sufficient; label quality differs across
sources. \textbf{Pipeline is faithful.} Re-running the shipped filter's
exact Mathlib dataset through the new pipeline reproduces $833/855$, so the
deficit is data composition, not a pipeline regression.

\textbf{Where the $+2$ lives.} A per-bucket breakdown shows the filter is
a near-exact no-op on the easy/medium $\approx\!79\%$ of the benchmark; all
movement is in the hard bucket. The $+2$ are three near-duplicate
square-root recurrence theorems minus one loss. A controlled synthetic
sweep of $120$ leak-free integer recurrences finds the filter exactly
inert ($0$ rescued, $0$ broken): scoring by \emph{lemma identity}, it
cannot distinguish wasteful from useful instances of the \emph{same}
hypothesis, so it helps only when the explosion is lemma-\emph{diverse}.
On the low-contention partition, paired repeated trials give the filter
$-4.9\%$ walltime (95\% CI $[-20.3,-16.0]$\,s) while solving $+2$ every
repetition.

\section{Split: Static-Model Details}
\label{app:split-models}
The four static-feature policies behind \cref{tab:rescue} are: a
gradient-boosted cost regressor; a generation-ordering rule (validated
held-out at $+9.5$ points optimal-pick on a cost metric); a doom/value
model trained on all rollouts in a {DAgger}-style
loop~\citep{ross2011dagger} (success-prediction AUC $0.984$ at the
state level); and a failure-aware success classifier trained on
${\sim}15$K forced-choice labels with candidate-level AUC $0.85$, all
evaluated by-decision held-out. They score the same goal- and
candidate-level features: the number of cases a split induces, whether it
is recursive, the source theory, the generation depth, the split depth,
and the number of \ematch{}ing rounds so far. Despite strong aggregate
metrics, every one is at or below uniform random on the rescuable
split-failures, because the discriminative signal (whether a split
explodes) is dynamic and not present in the static features.

\paragraph{Per-theorem behavior.}
\cref{tab:perthm} lists the seven decisive \textsc{numina} cases behind
the live cascade result of \cref{sec:split}: the five \grind{} timeouts
the cascade rescues, and the two goals where always-on lookahead regresses
but the cascade does not (it never runs lookahead on a goal stock already
solves). Rescue depth is the minimum number of further splits the rescuing
branch needs to close.

\begin{table}[h]
\centering\small
\begin{tabular}{lcccr}
\toprule
theorem & stock & lookahead & cascade & rescue depth \\
\midrule
000125 & timeout & solve   & solve & 9 \\
000281 & timeout & solve   & solve & 9 \\
000470 & timeout & solve   & solve & 5 \\
000530 & timeout & solve   & solve & 4 \\
000619 & timeout & solve   & solve & 3 \\
\midrule
000027 & solve   & timeout & solve & 57 \\
000220 & solve   & timeout & solve & 10 \\
\bottomrule
\end{tabular}
\caption{Per-theorem behavior on the seven decisive cases (anonymized
\textsc{numina} IDs). Top: the five stock timeouts the cascade rescues.
Bottom: the two goals where always-on lookahead regresses, which the
cascade leaves to stock and therefore still solves.}
\label{tab:perthm}
\end{table}

\section{Premise Augmentation}
\label{app:premise}
\begin{table}[h]
\centering\small
\begin{tabular}{lr}
\toprule
Outcome (n$=$1484, \textsc{numina}) & count \\
\midrule
base \grind{} solves            & 76 (5.1\%) \\
aug \grind{}$[$prem$]$ solves   & 83 (5.6\%) \\
\emph{net effect of premises}   & $+7$ (${\approx}$neutral) \\
\midrule
RESCUE (aug ok, base fail)      & \textbf{27 $=$ 1.9\% of base-failures} \\
REGRESSION (base ok, aug fail)  & \textbf{20 $=$ 26\% of base-solves} \\
\bottomrule
\end{tabular}
\caption{Premise-augmentation oracle on \textsc{numina}. Even
ground-truth premises rescue ${\sim}2\%$ of failures but break ${\sim}1/4$
of the (few) solves via \ematch{} blowup, netting ${\approx}0$.}
\label{tab:aug}
\end{table}

\noindent The additive lever is small and idea-gap-bound: most failures
lack a proof \emph{idea} (intermediate steps, constructions), not a
citable lemma---the ``idea gap'' familiar from hammer
evaluations~\citep{blanchette2016hammering,czajka2018coqhammer}---and the
premises that are added themselves trigger \ematch{} blowup. But a failure-triggered cascade (run plain \grind{};
only on failure retry with premises) banks the rescues with zero
regression by construction. Of the rescues, ${\approx}53\%$ need a single
premise and ${\approx}74\%$ need at most two, favoring retrieval over a
fixed-vocabulary classifier.

\end{document}